\documentclass[conference]{IEEEtran}
\IEEEoverridecommandlockouts
\usepackage{cite}
\usepackage{amsmath,amssymb,amsfonts}
\usepackage{algorithmic}
\usepackage{graphicx}
\usepackage{adjustbox}
\usepackage{booktabs}
\usepackage{tabularx}
\newcolumntype{Y}{>{\centering\arraybackslash}X}
\usepackage{textcomp}
\usepackage{xcolor}
\usepackage{comment}
\usepackage{subcaption}
\usepackage{hyperref}
\usepackage{flushend}
\usepackage{array}
\definecolor{mypink}{RGB}{210, 3, 141}

\def\BibTeX{{\rm B\kern-.05em{\sc i\kern-.025em b}\kern-.08em
    T\kern-.1667em\lower.7ex\hbox{E}\kern-.125emX}}
\begin{document}

\title{Squeezed Deep 6DoF Object Detection\\using Knowledge Distillation\\}

\author{\IEEEauthorblockN{
Heitor Felix\,$^{1}$, 
Walber M. Rodrigues\,$^{2}$,
David Mac\^edo\,$^{3,4}$,
Francisco Sim\~oes\,$^{1,5}$,\\
Adriano L. I. Oliveira\,$^{3}$,
Veronica Teichrieb\,$^{1}$, and
Cleber Zanchettin\,$^{3, 6}$
}
\IEEEauthorblockA{\,
$^{1}$Voxar Labs, Centro de Inform\'atica, Universidade Federal de Pernambuco, Recife, Brazil\\
$^{2}$Rob\^oCIn, Centro de Inform\'atica, Universidade Federal de Pernambuco, Recife, Brazil\\
$^{3}$Centro de Inform\'atica, Universidade Federal de Pernambuco, Recife, Brazil\\
$^{4}$Montreal Institute for Learning Algorithms, University of Montreal, Quebec, Canada\\
$^{5}$Instituto Federal de Pernambuco, Campus Belo Jardim, Belo Jardim, Brazil\\
$^{6}$Department of Chemical and Biological Engineering, Northwestern University, Evanston, United States of America\\
Emails: \{hcf2, wmr, dlm, fpms, alio, vt, cz\}@cin.ufpe.br}
}

\maketitle

\begin{abstract}
The detection of objects considering a 6DoF pose is a common requirement to build virtual and augmented reality applications. It is usually a complex task which requires real-time processing and high precision results for adequate user experience. Recently, different deep learning techniques have been proposed to detect objects in 6DoF in RGB images. However, they rely on high complexity networks, requiring a computational power that prevents them from working on mobile devices. In this paper, we propose an approach to reduce the complexity of 6DoF detection networks while maintaining accuracy. We used Knowledge Distillation to teach portables Convolutional Neural Networks (CNN) to learn from a real-time 6DoF detection CNN. The proposed method allows real-time applications using only RGB images while decreasing the hardware requirements. We used the LINEMOD dataset to evaluate the proposed method, and the experimental results show that the proposed method reduces the memory requirement by almost 99\% in comparison to the original architecture with the cost of reducing half the accuracy in one of the metrics. \textit{Code is available at \href{https://github.com/heitorcfelix/singleshot6Dpose}{\color{mypink}https://github.com/heitorcfelix/singleshot6Dpose}}.
\end{abstract}


\section{Introduction}

A key functionality in augmented reality applications is the ability to recover the pose of an object considering the position and orientation of the camera. Such information allows the creation of the interaction between the real and virtual environments. In robotics, this information allows the robots to interact, avoid collisions, and manipulate objects.

\begin{figure}[!h]
\includegraphics[width=\columnwidth]{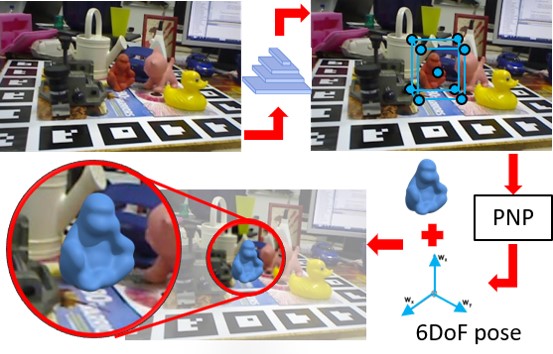}
\caption{Flowchart of our method of 6DoF object detection and its possible application in augmented reality. The squeezed network receives an input image, estimates the bounding box of the object to be detected. Then, its 6DoF pose is estimated using a PnP algorithm. With the pose, it is possible to create augmented reality applications by rendering the 3D model over the image.}
\label{fig-focus}
\end{figure}

The object detection with two degrees of freedom retrieves the object's coordinates relative to the image plane. In object detection with three degrees of freedom, the method recovers the position or translation of a real-world object corresponding to the camera. These points have the information about the depth, but not the rotation. In object detection with six degrees of freedom (6DoF), there are three degrees for object translation and three degrees for object rotation considering the relation to the camera. Hence, with the 6DoF approach, it is possible to make robots to pick an object precisely \cite{hernandez2016robotic} or to create augmented reality applications with more interaction between real and virtual objects \cite{tan20176d}. Approaches to this problem often use an array of sensors instead of an RBG camera \cite{tan20176d}.

There are many efficient methods for detecting objects in 6DoF. However, these methods suffer from challenges as weakly textured or untextured objects, and low-resolution images \cite{tjaden2017real}. Among the relevant works for 6DoF object detection, Tekin et al. \cite{tekin2018real} presents results for pose estimation using only RGB images as input and achieved real-time detection for untextured objects using a YOLOv2-based architecture \cite{redmon2017YOLO9000}. Many works have already improved the architecture of YOLOv2 \cite{huang2018yolo, mehta2018object, redmon2018yolov3}, and one of these improvements can be efficient for Tekin et al. architecture. The evaluation of the method considers the object reprojection and pose errors from the object model points in camera coordinates analyzed in the LINEMOD dataset. The inference time and network size are also metrics compared to the related works. These evaluation metrics are also used in \cite{rad2017bb8, brachmann2016bb82, kehl2017ssd6d, zakharov2019dpod}.

The Tekin et al. \cite{tekin2018real} approach achieves a reasonable pose estimate in real-time but requires high-end hardware. This limitation makes embedding the method on mobile devices and robots a challenging task, making this approach less accessible for augmented reality and robotics applications.

There are different approaches to optimize deep learning networks \cite{cheng2017survey, hinton2015distilling, han2015learning, he2017channel}. An approach that has been used recently is the Knowledge Distillation (KD) \cite{hinton2015distilling}. The main idea of this technique is to use a larger and complex network with high-quality prediction results as a target model (teacher). Consequently, the small network (student) is trained with the help of the teacher. The idea is to train the student network targeting the teacher's network knowledge instead of using only the original problem labels, which are, in general, more complex and challenging to learn in relation to the teacher's assistance.

In Mehta and Ozturk's work \cite{mehta2018object}, the authors used Knowledge Distillation (KD) for object detection. The student network is the Tiny-YOLO network, and the YOLOv2 network is the teacher network. Mehta and Ozturk achieved better accuracy with the same network using KD training in relation to Tiny-YOLO without KD. Since the network used for Mehta and Ozturk is a direct reduction of the network used as a basis by Tekin et al., the results of Mehta and Ozturk show the potential performance gain of using KD to 6DoF object detection.

\begin{figure}
\centering
\includegraphics[width=\columnwidth]{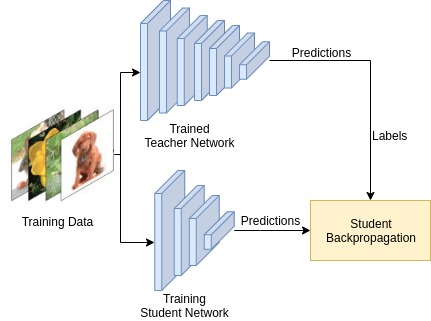}
\caption{The Knowledge Distillation used in this work changes the training labels of the student network by the prediction of the teacher network.}
\label{fig:knowledge_distillation}
\end{figure}

In this paper, we use the simplest KD technique that replaces only target labels of a student training to optimize adapted Tekin et al. architecture for Deep 6DoF Object Detection. This KD only replaces the label during the training of the student network with the prediction of the teacher network. The replacement allows the student network to learn from a network that already has its activation regions defined and makes learning easier. Our 6DoF object detection method is summarized in Figure \ref{fig-focus} and the process diagram of it is shown in Figure \ref{fig:knowledge_distillation}. We performed experiments with the LINEMOD dataset \cite{hinterstoisser2012linemod} and compared it with the original model.

This paper is organized as follows. In Section \ref{related_works}, we revise works related to complexity reduction in Convolution Neural Networks (CNNs). In Section, \ref{methodology}, we describe our methodology using KD for CNNs. In Section \ref{experiments}, we define our experimental protocol, including Dataset, Evaluation Metrics, and the Knowledge Distillation for pose estimation. Section \ref{main} shows the obtained results and, finally, Section \ref{conclusion} gives some concluding remarks.

\section{Related Works}
\label{related_works}

Several techniques have been proposed to reduce the complexity of deep networks, including Parameter Pruning and Sharing, Low-rank Factorization, Transferred or Compact Convolution Filters, and Knowledge Distillation (KD) \cite{cheng2017survey}.

Pruning weights was one of the primary methods for reducing complexity \cite{lecun1990optimal}. When first proposed, parameter pruning removed meaningless neurons connections. Although this method can drastically reduce the complexity of fully-connected architectures, in CNN networks, this approach is overwhelmed by the number of weights in complex networks. This problem was solved by He et al. \cite{he2017channel}, which introduced a channel-pruning to remove a filter altogether.

Low-rank Factorization uses the intrinsic CNN characteristic of matrix multiplication to create filters that can be separated into two matrices multiplications. This method intends to reduce the multiplications needed to apply a filter and can achieve up to two times speed-up on complex networks such as VGG-16 \cite{cheng2017survey}.

Compact Convolution Filters is another popular and powerful method. Based on changing complex 7x7 and 5x5 filters by 3x3 or 1x1 filters to reduce the number of operations, it is gaining attention in the field. Works such as \cite{huang2018yolo} and \cite{mehta2018object} broadly uses that method to reduce complexity without losing too much accuracy.

Another method of optimization classified is Knowledge Distillation (KD) \cite{cheng2017survey}. This method proposes that a complex and powerful network teaches a smaller network, which is trained using the output of the bigger network by trying to mimic it. This method allows the smaller network to achieve better results by learning from the teacher network than trained with the original data \cite{mehta2018object}. It proved to be efficient, and some variations were proposed for the most diverse problems \cite{tang2019bert}.

Basic KD is done only by replacing the student network label with the teacher network output. Some works vary the use of KD. In \cite{mehta2018object}, different loss calculations are used, taking into account the difference between the student network and the teacher network outputs.
In \cite{tang2019bert, radosavovic2018data}, KD is used for the teacher network to generate new data for the training of the student network. In this work, only the basic KD was used, leaving its changes for possible future works.

Our work aims to create a faster and lighter version of an object detection model based on Tekin et al. \cite{tekin2018real} to estimate the pose of an object. Also, use the original architecture to train our squeezed architecture using KD. Thus, improve the inference time and memory consumption with acceptable accuracy reductions.

\begin{figure*}
\centering
\begin{tabular}{c}
\subcaptionbox{}{\includegraphics[width=00.99\columnwidth]{Figures/linemod.jpg}}%
\hspace{0.1cm}
\subcaptionbox{}{\includegraphics[width=0.99\columnwidth]{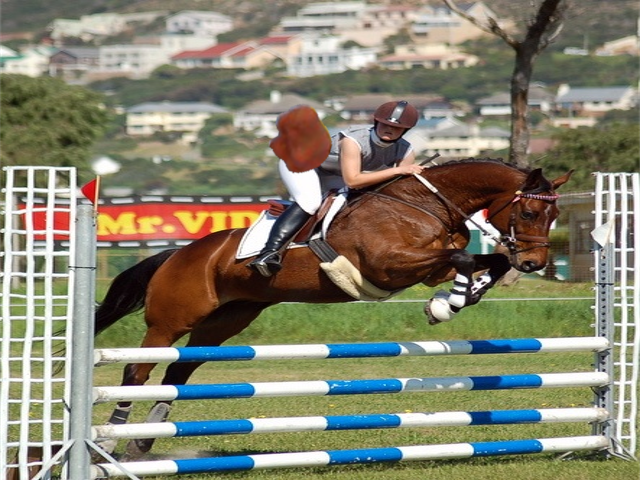}}
\\
\end{tabular}
\caption{Data augmentation in the LINEMOD dataset for network training. (a) An original image of the LINEMOD dataset. (b) The Ape object added to a random background of the VOC2012 dataset with a small rotation and dislocation of the object.}
\label{fig:data_augmentation}
\end{figure*}

\section{Knowledge Distilled Squeezed Models}
\label{methodology}

In 6DoF detection problems, KD approaches can be particularly useful because the larger network learned a smoother decision region than what can be learned directly from the dataset. So the smaller network, called student network, does not have to learn hard instances during training, because the larger network, called teacher network, has already dealt with these cases, creating a specific activation region for the problem. As expected, in KD, the student network does not achieve the accuracy of the teacher network, since the teacher network is more complex and has more parameters, thus requiring more computational power than the student network.

In this work, Tekin et al. \cite{tekin2018real} propose an architecture using YOLOv2 \cite{redmon2017YOLO9000} as its base and by changing its last layer and adding a second shorter network path. The shorter path filters have a size different from the first path, forcing the network to learn a completely different set of filters. Problems with a multidimensional activation region, such as inpainting \cite{wang2018image}, use this path addition.

In our work, we use a modified architecture of Tekin et al. to reduce network complexity. With this new architecture, we use the KD represented in Figure \ref{fig:knowledge_distillation}. In this work, to student network, we substitute the basis of the original work that is the YOLOv2 architecture by an architecture based on YOLO-LITE \cite{huang2018yolo} architecture.

The YOLO-LITE work was used for its focus on small architectures for computers without GPU. With the good performance obtained by these networks, they are good options for the basis of the 6DoF object detection architecture.

In YOLO-LITE, several architectures were proposed. These architectures are called Trials and range from 1 to 13. We chose the best Trial based on the trade-off between complexity and mAP. According to Huang et al. \cite{huang2018yolo}, Trial 3 without batch normalization (BN) got the best result. Since the input image size in Trial 3 is 224x224 pixels and the architecture used as teacher and base to student uses imagens with 416x416 pixels, we will use Trial 10 as a basis for the squeezed network, using with and without BN.

\begin{table*}
    \centering
    \begin{tabularx}{\textwidth}{lYYYYY}
    \toprule
    Object & \multicolumn{2}{c}{Distilled Models}       & \multicolumn{2}{c}{Non-distilled Models}   \\
    \cmidrule(lr){2-3} \cmidrule(lr){4-5}
       & YOLO-LITE Trial 10 & YOLO-LITE Trial 10 w/ BN & YOLO-LITE Trial 10 & YOLO-LITE Trial 10 w/ BN & Tekin et al \cite{tekin2018real} \\
    
    \midrule
    
    Ape             & \textbf{51.428}   & 46.761            & 44.095            & 45.809            & 92.100 \\
    Benchvise       & 45.736            & 45.445            & \textbf{46.608}   & 40.406            & 95.060 \\
    Cam             & 36.176            & 27.156            & \textbf{36.568}   & 26.666            & 93.240 \\
    Can             & 52.854            & 51.181            & \textbf{53.444}   & 48.425            & 97.440 \\
    Cat             & 43.712            & 40.718            & 38.722            & \textbf{46.906}   & 97.410 \\
    Driller         & 23.984            & 20.317            & \textbf{31.714}   & 16.253            & 79.410 \\
    Duck            & 30.140            & 24.600            & \textbf{32.394}   & 29.577            & 94.650 \\
    Eggbox          & 38.591            & 37.652            & 38.967            & \textbf{42.065}   & 90.330 \\
    Glue            & \textbf{50.965}   & 46.718            & 50.482            & 45.173            & 96.530 \\
    Holepuncher     & \textbf{36.822}   & 25.784            & 34.443            & 30.542            & 92.860 \\
    Iron            & \textbf{25.638}   & 24.821            & 24.821            & 25.331            & 82.940 \\
    Lamp            & 29.654            & 24.856            & \textbf{30.134}   & 28.598            & 76.870 \\
    Phone           & \textbf{39.673}   & 13.256            & 34.774            & 27.570            & 86.070 \\
    \midrule
    Average         & \textbf{38.875}   & 33.021            & 38.244            & 34.871            & 90,370 \\
    \bottomrule
    
    \end{tabularx}
	\caption{Results of 2D Projection for Distilled and non-distilled YOLO-LITE Trial 10, YOLO-LITE Trial 10 with Batch Normalization (BN), and the teacher network by Tekin et al. \cite{tekin2018real}. In bold, the best result for the lite models per object.}
	\label{table:2d_complete} 
\end{table*}
\begin{table*}
    \centering
    \begin{tabularx}{\textwidth}{lYYYYY}
    \toprule
    Object & \multicolumn{2}{c}{Distilled Models}       & \multicolumn{2}{c}{Non-distilled Models}   \\
    \cmidrule(lr){2-3} \cmidrule(lr){4-5}
       & YOLO-LITE Trial 10 & YOLO-LITE Trial 10 w/ BN & YOLO-LITE Trial 10 & YOLO-LITE Trial 10 w/ BN & Tekin et al \cite{tekin2018real} \\
    
    \midrule
    
    Ape             & 4.095             & 4.380            & 2.571             & \textbf{5.619}    & 21.620 \\
    Benchvise       & 30.620            & \textbf{30.813}  & 30.523            & 29.360            & 81.800 \\
    Cam             & 12.156            & 10.294           & \textbf{14.705}   & 10.392            & 36.570 \\
    Can             & 22.637            & 18.503           & \textbf{23.228}   & 21.161            & 68.800 \\
    Cat             & 7.185             & 9.880            & 7.185             & \textbf{10.479}   & 41.820 \\
    Driller         & 21.110            & 19.127           & \textbf{25.173}   & 13.974            & 63.510 \\
    Duck            & 3.192             & 2.629            & \textbf{3.380}    & 2.723             & 27.230 \\
    Eggbox          & 10.328            & 9.295            & 12.582            & \textbf{12.957}   & 69.580 \\
    Glue            & 11.389            & 10.135           & 9.845             & \textbf{11.583}   & 80.020 \\
    Holepuncher     & \textbf{9.419}    & 7.326            & 8.087             & 7.897             & 42.630 \\
    Iron            & 25.331            & 27.885           & \textbf{28.804}   & 27.374            & 74.970 \\
    Lamp            & \textbf{25.815}   & 21.305           & 24.088            & 23.416            & 71.110 \\
    Phone           & \textbf{20.365}   & 11.623           & 19.500            & 15.177            & 47.740 \\
    \midrule
    Average         & 15.665            & 14.298           & \textbf{16.129}   & 14.778            & 55.950 \\
    \bottomrule
    
    \end{tabularx}
	\caption{Results of 3D Transformation for Distilled and non-distilled YOLO-LITE Trial 10, YOLO-LITE Trial 10 with Batch Normalization (BN), and the teacher network by Tekin et al. \cite{tekin2018real}. In bold, the best result for the lite models per object.}
	\label{table:m3d_complete} 
\end{table*}

Our model is trained by modifying the label of the associated image. Instead of using the dataset label, we employ the output of the teacher architecture. In this case, it is the original Tekin et al. \cite{tekin2018real} architecture, associated with the image of the dataset used. All other training steps of the original network were maintained.

The baseline model has a complex architecture composed of 23 layers with filters 1x1 and 3x3 filter sizes. Although having small filter sizes, it uses up to 1024 filters per layer, highly increasing the network size.

\begin{figure*}
\centering
\begin{subfigure}[b]{0.99\columnwidth}
\includegraphics[width=0.99\columnwidth]{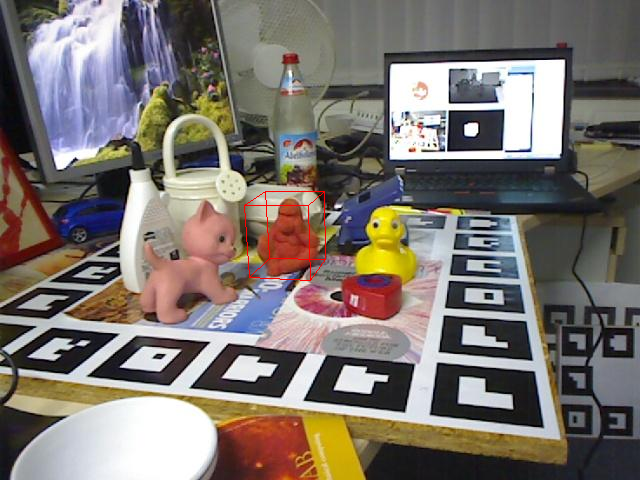}
\end{subfigure}
\begin{subfigure}[b]{0.99\columnwidth}
\includegraphics[width=0.99\columnwidth]{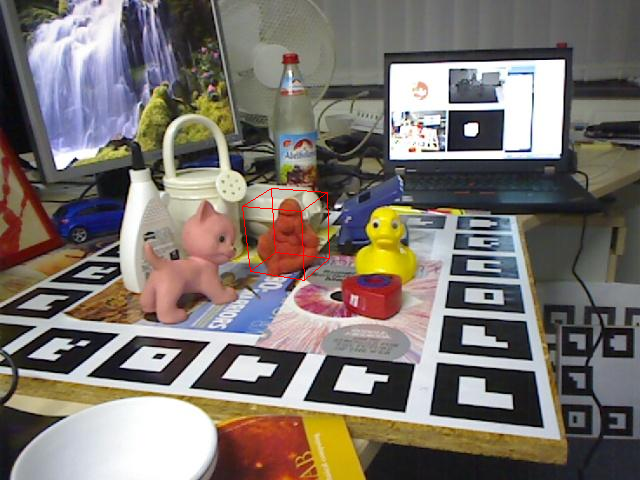}
\end{subfigure}
\par\medskip
\begin{subfigure}[b]{0.99\columnwidth}
\includegraphics[width=0.99\columnwidth]{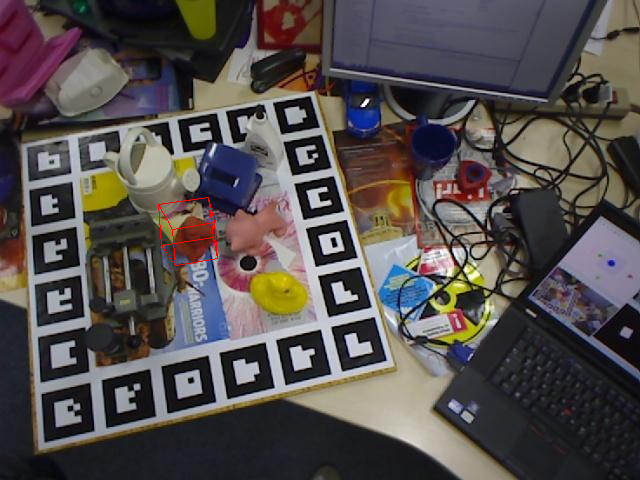}
\end{subfigure}
\begin{subfigure}[b]{0.99\columnwidth}
\includegraphics[width=0.99\columnwidth]{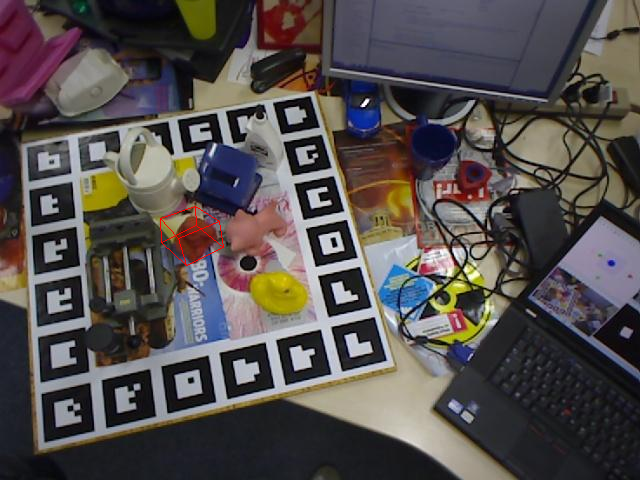}
\end{subfigure}
\par\medskip
\begin{subfigure}[b]{0.99\columnwidth}
\includegraphics[width=0.99\columnwidth]{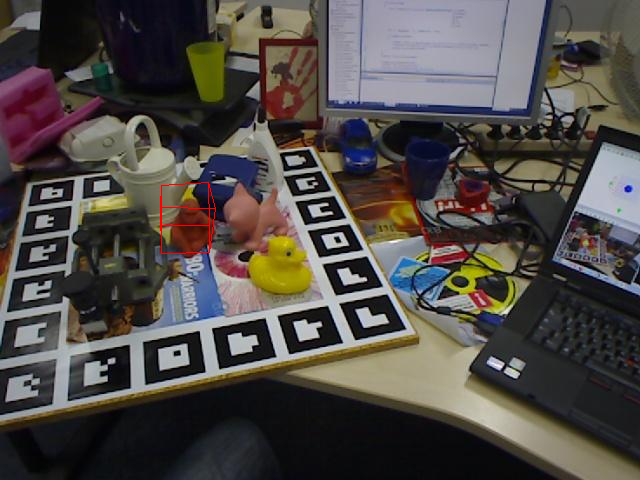}
\end{subfigure}
\begin{subfigure}[b]{0.99\columnwidth}
\includegraphics[width=0.99\columnwidth]{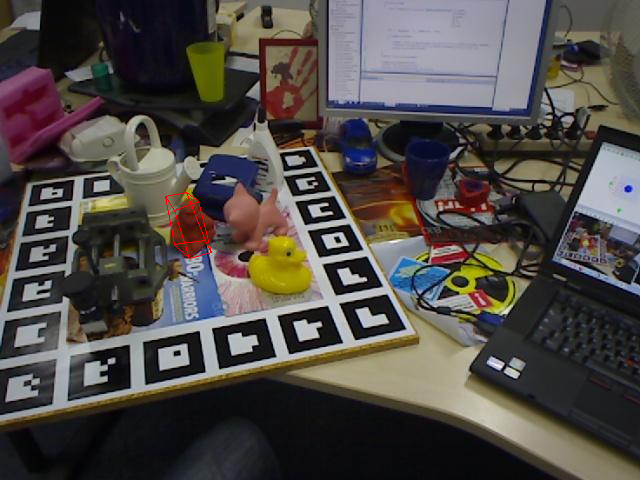}
\end{subfigure}
\caption{Qualitative Results on object Ape. On the left, detection performed with the original Tekin et al. network. On the right, detection using YOLO-LITE Trial 10 without BN.}
\label{fig:quality}
\end{figure*}

\section{Experiments}
\label{experiments}

All experiments were performed using the Ubuntu 18.04 operating system and developed with Python 3.7 using the Deep Learning API PyTorch 1.2. We used a computer with an Intel Xeon 1.7GHz x 16 processor, 16GB RAM, and GeForce RTX 2080Ti GPU.

Initially, in Subsection \ref{dataset}, the dataset used is explained. In Subsection \ref{metrics}, the metrics used are detailed. The details of the experiments performed are in Section \ref{main}.

\subsection{Dataset}
\label{dataset}
The dataset used is the LINEMOD \cite{hinterstoisser2012linemod}, which is also used in other works \cite{tekin2018real, rad2017bb8, brachmann2016bb82, kehl2017ssd6d} for a reliable comparison. Also, LINEMOD is one of the main datasets used for 6DoF object detection, having several challenges such as occlusion, polluted background, blurring, and lighting change. Figure \ref{fig:data_augmentation} (a) is one of the images contained in the LINEMOD Dataset.

LINEMOD is made up of thirteen objects, where each object has about twelve hundred color images in 640x480 JPEG format and a 3D model. Also, for each image, the binary mask for the objects and their rotation and translation vector values are informed.

Training images went through a process of synthetic data augmentation as used by Tekin et al. During synthetic data augmentation, new images are created for each image for the object being trained. The object is segmented using its binary mask, available in LINEMOD, and added to a random background from the dataset VOC2012 \cite{everingham2010voc}, besides that, the object in the image is rotated and translated aiming to avoid overfitting and to increase the number of images available for training. Figure \ref{fig:data_augmentation} (b) was obtained from the Ape object data augmentation.

\subsection{Evaluation Metrics}
\label{metrics}

For performance evaluation, we used two evaluation metrics for 6DoF pose estimation and the pose inference time. These metrics were also used in related works as \cite{tekin2018real, rad2017bb8, brachmann2016bb82, kehl2017ssd6d, zakharov2019dpod} and evaluate the prediction of the pose of objects. The metrics used are the 2D Projection \cite{brachmann2016bb82} and 3D Transformation \cite{hinterstoisser2012linemod}.

The 2D Projection metric is defined as the average of 2D reprojected object hit on the image from the test set images. The reprojection hit is calculated using \eqref{R2D2} when the projection distance is lower or equal than five pixels is considered a hit \cite{brachmann2016bb82}.

\begin{equation}
\Delta_{2D}(x_1,x_2) = 
    \begin{cases} 1, & if \ | x_1 - x_2 | \leq 5 \ pixels \\ 0, & otherwise 
    \end{cases}
\label{R2D2}
\end{equation}

The 3D Transformation metric is defined as the average of object model transformation hit on the image from the test set images. Projection hit is calculated using the 3D model points of the object, it is ground truth matrix of extrinsic parameters, and its estimated matrix of extrinsic parameters. For each point of the transformed 3D model using the matrices, the pose error is calculated from a distance between the transformed point using the ground truth matrix and the transformed point using the estimated matrix. The pose error is calculated using \eqref{eqPose6d}, where $ \mathit{n} $ is the total number of points in the 3D model of the object. Object model transformation hit is considered if $ \mathit{m} $ is lower or equal than 10\% of object model diameter \cite{hinterstoisser2012linemod}. The 3D Transformation metric consists of this hit average in the test images.

\begin{equation}
m = \frac{1}{n}\sum^n_{i=1}\left \| (r_{gt}x_i + t_{gt}) - (r_{pr}x_i + t_{pr})\right \|
\label{eqPose6d}
\end{equation}

\section{Results and Discussion}
\label{main}

According to the mentioned models, for each metric, the values shown in Tables \ref{table:2d_complete}, \ref{table:m3d_complete}, \ref{table:time} and \ref{table:model_sizes} were obtained. The use of the 3D Transformation metric produces the results shown in Table \ref{table:m3d_complete}. Here we can see that for all architectures, the non-distilled models barely outperforms the distilling ones. For the 2D Reprojection technique, however, as shown in Table \ref{table:2d_complete}, the results are balanced between distilled and non-distilled models. A detailed result for each object in the dataset is showed at Table \ref{table:2d_complete} and Table \ref{table:m3d_complete}.

As an explanation, the results show that distilling can improve the performance in this application. 3D Transformation is a hard task to be performed, as the baseline model only achieves 55.95\%. This result tells us that the quality of the target in which the student model learns plays an important role in its performance. On the other side, as 2D projection is a simpler task, distilling and non-distilled models have performance equally distributed.

The negative performance of distilled models is explained by Saputra et al. \cite{saputra2019distilling}. In their work, it is proved that if the teacher is not reliable enough, the KD will not improve accuracy as it will add noise to the training.

For the inference time, shown in Table \ref{table:time}, it can be observed the significant gain compared to the teacher network because it is much more complex, requiring longer processing time. The student network achieves an inference time more than 3x faster.

\begin{table}
\centering 
    \renewcommand{\arraystretch}{1.25} 
    \begin{tabularx}{\columnwidth}{lYY}
        \toprule 
            Model & Inference Time (ms)  \\ 
        \midrule
            Baseline & 6.5 \\
            YOLO-LITE Trial 10 w/ BN & \textbf{2.0} \\
            YOLO-LITE Trial 10 & \textbf{2.0} \\
        \bottomrule
    \end{tabularx} 
\caption{Evaluation of inference time in the Ape object. In bold, the best values per method.}
\label{table:time} 
\end{table}

The network weight file sizes were compared to analyze the differences in complexity between network architectures. The sizes of networks in MB are shown in Table \ref{table:model_sizes}. In Tables \ref{table:time} and \ref{table:model_sizes}, Trial 10 with BN and without BN have the same values because of its shares the same architecture.
The big difference between the weight sizes of the network directly affects the memory space to use the networks. It happens due to the difference in the number of parameters of the network architecture. The table shows that the architecture based on YOLO-LITE is almost 99\% smaller than the original reference architecture.

From a visual qualitative point of view, Figure \ref{fig:quality} shows that our Distilled YOLO-LITE Trial 10 without batch normalization proposal produces similar results when compared with the baseline approach. It also achieves a massive gain in time and processing power. Our results open new possibilities of using the remaining processing time to increase its precision by integrating a refinement technique as the DeepIM\cite{Li_2019}.

\begin{table}
\centering 
	\renewcommand{\arraystretch}{1.25} 
	\begin{tabularx}{\columnwidth}{lYY}

		\toprule 
		Model & Weights (MB)  \\ 
		\midrule
		Baseline & 202.3 \\
		Yolo-Lite Trial 10 w/ BN & \textbf{2.2} \\
		Yolo-Lite Trial 10 & \textbf{2.2} \\
		\bottomrule
	\end{tabularx} 
	\caption{Sizes in MB for each evaluated model. In bold, the smallest models.}
	\label{table:model_sizes} 
\end{table}

\section{Conclusions and Future Works}
\label{conclusion}

This work showed the benefits and consequences of exchanging the basis network for the 6DoF object detection technique proposed by Tekin et al. \cite{tekin2018real} for the YOLO-LITE  architecture \cite{huang2018yolo}. Thus, it is possible to analyze and compare which network to use for 6DoF object detection. Thus choose the right architecture for the target hardware, which can be a smartphone or a robot. It also shows the effects of a basic KD \cite{hinton2015distilling} usage to train a simpler network with YOLO-LITE \cite{huang2018yolo} basis. This work's main goal was to show how it is possible to reduce the 6DoF object detection network proposed by Tekin et al. by more than 90x (weights in MB), reducing the hit rate by a smaller factor, but still achieving comparable results in the qualitative evaluation.

In robust computer vision systems, detection and tracking techniques are usually combined. The detection is used to obtain the initial pose, and then the tracking is used frame by frame. When the tracking accumulates errors or fails, the detection is called up again for a system reboot.

Many of these systems look for lightness to run on embedded systems or mobile devices. For these systems, it would be necessary to use lightweight detectors. Also, even for systems that use powerful hardware, it can be advantageous to use lightweight detectors since it would be used only a few times during an application. Thus, even reducing the total accuracy, the network developed in this work would be an option for different systems.

The methodology proposed by this work can also be used in future works where new state-of-the-art 6DoF object detection techniques like DPOD \cite{zakharov2019dpod} can be used as a new teacher network. This assumption is supported by Saputra et al. \cite{saputra2019distilling}, which claims that KD achieves greater can be used as a new teacher network and thus improve the efficiency of the use of KD that achieves greater efficiency with detectors that generate less noise.

Test other KD versions based on mentioned techniques \cite{mehta2018object, saputra2019distilling, tang2019bert} can improve KD usage. These works use different loss calculation based on the difference between student and teacher outputs \cite{mehta2018object}. Also, Tang et al. \cite{tang2019bert} and Radosavovic et al. \cite{radosavovic2018data} also use KD to generate new data for student network training.

\bibliographystyle{IEEEtran}
\bibliography{IEEEabrv,thebibliography}

\begin{thebibliography}{10}
\providecommand{\url}[1]{#1}
\csname url@samestyle\endcsname
\providecommand{\newblock}{\relax}
\providecommand{\bibinfo}[2]{#2}
\providecommand{\BIBentrySTDinterwordspacing}{\spaceskip=0pt\relax}
\providecommand{\BIBentryALTinterwordstretchfactor}{4}
\providecommand{\BIBentryALTinterwordspacing}{\spaceskip=\fontdimen2\font plus
\BIBentryALTinterwordstretchfactor\fontdimen3\font minus
  \fontdimen4\font\relax}
\providecommand{\BIBforeignlanguage}[2]{{%
\expandafter\ifx\csname l@#1\endcsname\relax
\typeout{** WARNING: IEEEtran.bst: No hyphenation pattern has been}%
\typeout{** loaded for the language `#1'. Using the pattern for}%
\typeout{** the default language instead.}%
\else
\language=\csname l@#1\endcsname
\fi
#2}}
\providecommand{\BIBdecl}{\relax}
\BIBdecl

\bibitem{hernandez2016robotic}
C.~Hernandez, M.~Bharatheesha, W.~Ko, H.~Gaiser, J.~Tan, K.~van Deurzen,
  M.~de~Vries, B.~Van~Mil, J.~van Egmond, R.~Burger \emph{et~al.}, ``Team
  delft’s robot winner of the amazon picking challenge 2016,'' in \emph{Robot
  World Cup}.\hskip 1em plus 0.5em minus 0.4em\relax Springer, 2016, pp.
  613--624.

\bibitem{tan20176d}
D.~J. Tan, N.~Navab, and F.~Tombari, ``6d object pose estimation with depth
  images: A seamless approach for robotic interaction and augmented reality,''
  \emph{arXiv preprint arXiv:1709.01459}, 2017.

\bibitem{tjaden2017real}
H.~Tjaden, U.~Schwanecke, and E.~Schomer, ``Real-time monocular pose estimation
  of 3d objects using temporally consistent local color histograms,'' in
  \emph{Proceedings of the IEEE International Conference on Computer Vision},
  2017, pp. 124--132.

\bibitem{tekin2018real}
B.~Tekin, S.~N. Sinha, and P.~Fua, ``Real-time seamless single shot 6d object
  pose prediction,'' in \emph{Proceedings of the IEEE Conference on Computer
  Vision and Pattern Recognition}, 2018, pp. 292--301.

\bibitem{redmon2017YOLO9000}
J.~Redmon and A.~Farhadi, ``Yolo9000: better, faster, stronger,'' in
  \emph{Proceedings of the IEEE conference on computer vision and pattern
  recognition}, 2017, pp. 7263--7271.

\bibitem{huang2018yolo}
R.~Huang, J.~Pedoeem, and C.~Chen, ``Yolo-lite: a real-time object detection
  algorithm optimized for non-gpu computers,'' in \emph{2018 IEEE International
  Conference on Big Data (Big Data)}.\hskip 1em plus 0.5em minus 0.4em\relax
  IEEE, 2018, pp. 2503--2510.

\bibitem{mehta2018object}
R.~Mehta and C.~Ozturk, ``Object detection at 200 frames per second,'' in
  \emph{Proceedings of the European Conference on Computer Vision (ECCV)},
  2018, pp. 0--0.

\bibitem{redmon2018yolov3}
J.~Redmon and A.~Farhadi, ``Yolov3: An incremental improvement,'' \emph{arXiv
  preprint arXiv:1804.02767}, 2018.

\bibitem{rad2017bb8}
M.~Rad and V.~Lepetit, ``Bb8: A scalable, accurate, robust to partial occlusion
  method for predicting the 3d poses of challenging objects without using
  depth,'' in \emph{Proceedings of the IEEE International Conference on
  Computer Vision}, 2017, pp. 3828--3836.

\bibitem{brachmann2016bb82}
E.~Brachmann, F.~Michel, A.~Krull, M.~Ying~Yang, S.~Gumhold \emph{et~al.},
  ``Uncertainty-driven 6d pose estimation of objects and scenes from a single
  rgb image,'' in \emph{Proceedings of the IEEE Conference on Computer Vision
  and Pattern Recognition}, 2016, pp. 3364--3372.

\bibitem{kehl2017ssd6d}
W.~Kehl, F.~Manhardt, F.~Tombari, S.~Ilic, and N.~Navab, ``Ssd-6d: Making
  rgb-based 3d detection and 6d pose estimation great again,'' in
  \emph{Proceedings of the IEEE International Conference on Computer Vision},
  2017, pp. 1521--1529.

\bibitem{zakharov2019dpod}
S.~Zakharov, I.~Shugurov, and S.~Ilic, ``Dpod: 6d pose object detector and
  refiner,'' in \emph{Proceedings of the IEEE International Conference on
  Computer Vision}, 2019, pp. 1941--1950.

\bibitem{cheng2017survey}
Y.~Cheng, D.~Wang, P.~Zhou, and T.~Zhang, ``A survey of model compression and
  acceleration for deep neural networks,'' \emph{arXiv preprint
  arXiv:1710.09282}, 2017.

\bibitem{hinton2015distilling}
G.~Hinton, O.~Vinyals, and J.~Dean, ``Distilling the knowledge in a neural
  network,'' \emph{arXiv preprint arXiv:1503.02531}, 2015.

\bibitem{han2015learning}
S.~Han, J.~Pool, J.~Tran, and W.~Dally, ``Learning both weights and connections
  for efficient neural network,'' in \emph{Advances in neural information
  processing systems}, 2015, pp. 1135--1143.

\bibitem{he2017channel}
Y.~He, X.~Zhang, and J.~Sun, ``Channel pruning for accelerating very deep
  neural networks,'' in \emph{Proceedings of the IEEE International Conference
  on Computer Vision}, 2017, pp. 1389--1397.

\bibitem{hinterstoisser2012linemod}
S.~Hinterstoisser, V.~Lepetit, S.~Ilic, S.~Holzer, G.~Bradski, K.~Konolige, and
  N.~Navab, ``Model based training, detection and pose estimation of
  texture-less 3d objects in heavily cluttered scenes,'' in \emph{Asian
  conference on computer vision}.\hskip 1em plus 0.5em minus 0.4em\relax
  Springer, 2012, pp. 548--562.

\bibitem{lecun1990optimal}
Y.~LeCun, J.~S. Denker, and S.~A. Solla, ``Optimal brain damage,'' in
  \emph{Advances in neural information processing systems}, 1990, pp. 598--605.

\bibitem{tang2019bert}
R.~Tang, Y.~Lu, L.~Liu, L.~Mou, O.~Vechtomova, and J.~Lin, ``Distilling
  task-specific knowledge from bert into simple neural networks,'' \emph{arXiv
  preprint arXiv:1903.12136}, 2019.

\bibitem{radosavovic2018data}
I.~Radosavovic, P.~Doll{\'a}r, R.~Girshick, G.~Gkioxari, and K.~He, ``Data
  distillation: Towards omni-supervised learning,'' in \emph{Proceedings of the
  IEEE Conference on Computer Vision and Pattern Recognition}, 2018, pp.
  4119--4128.

\bibitem{wang2018image}
Y.~Wang, X.~Tao, X.~Qi, X.~Shen, and J.~Jia, ``Image inpainting via generative
  multi-column convolutional neural networks,'' in \emph{Advances in Neural
  Information Processing Systems}, 2018, pp. 331--340.

\bibitem{everingham2010voc}
M.~Everingham, L.~Van~Gool, C.~K. Williams, J.~Winn, and A.~Zisserman, ``The
  pascal visual object classes (voc) challenge,'' \emph{International journal
  of computer vision}, vol.~88, no.~2, pp. 303--338, 2010.

\bibitem{saputra2019distilling}
M.~R.~U. Saputra, P.~P. de~Gusmao, Y.~Almalioglu, A.~Markham, and N.~Trigoni,
  ``Distilling knowledge from a deep pose regressor network,'' in
  \emph{Proceedings of the IEEE International Conference on Computer Vision},
  2019, pp. 263--272.

\bibitem{Li_2019}
\BIBentryALTinterwordspacing
Y.~Li, G.~Wang, X.~Ji, Y.~Xiang, and D.~Fox, ``Deepim: Deep iterative matching
  for 6d pose estimation,'' \emph{International Journal of Computer Vision},
  Nov 2019. [Online]. Available:
  \url{http://dx.doi.org/10.1007/s11263-019-01250-9}
\BIBentrySTDinterwordspacing

\end{thebibliography}

\end{document}